\documentclass[10pt,twocolumn,letterpaper]{article}

\usepackage[pagenumbers]{cvpr} %

\usepackage[dvipsnames]{xcolor}
\newcommand{\red}[1]{{\color{red}#1}}

\usepackage{epsfig}
\usepackage{graphicx}
\usepackage{amsmath}
\usepackage{amssymb}
\usepackage{booktabs}
\usepackage{lib}
\usepackage{mathtools}

\usepackage{xspace}
\usepackage[accsupp]{axessibility}  %

\makeatletter
\DeclareRobustCommand\onedot{\futurelet\@let@token\@onedot}
\def\@onedot{\ifx\@let@token.\else.\null\fi\xspace}

\def\eg{\emph{e.g}\onedot} 
\def\ie{\emph{i.e}\onedot}

\usepackage{stmaryrd} %
\usepackage{paralist} %
\usepackage[symbol]{footmisc}

\definecolor{cvprblue}{rgb}{0.21,0.49,0.74}
\usepackage[pagebackref,breaklinks,colorlinks,citecolor=cvprblue]{hyperref}

\usepackage{color}
\definecolor{crimson}{rgb}{0.86, 0.08, 0.24}
\definecolor{gray}{rgb}{0.5,0.5,0.5}
\definecolor{green}{rgb}{0, 0.4, 0}
\definecolor{orange}{rgb}{1, 0.5, 0}
\definecolor{mahogany}{rgb}{0.75, 0.25, 0.0}
\definecolor{purple}{rgb}{0.6, 0, 0.6}
\definecolor{darkgreen}{rgb}{0, 0.4, 0}
\definecolor{frenchblue}{rgb}{0.0, 0.45, 0.73}
\definecolor{blue}{rgb}{0.0, 0.0, 0.65}
\definecolor{red}{rgb}{1,0,0}
\definecolor{yellow}{rgb}{1,1,0}
\definecolor{magenta}{rgb}{1,0,1}
\definecolor{pink}{rgb}{1,0.412,0.706}
\definecolor{newgreen}{rgb}{0, 0.6, 0.2}

\newcommand\mydots{\hbox to 1em{.\hss.\hss.}}

\newlength\paramargin
\newlength\figmargin
\newlength\subfigmargin
\newlength\subsecmargin
\newlength\tabmargin
\newlength\eqmargin

\newlength\presecmargin
\newlength\secmargin

\newlength\rulelength
\newcommand{\comment}[1]{}

\ifx \submission \undefined

    \newcommand {\hubert}[1]{{\color{violet}\textbf{Hubert: }#1}\normalfont}
    \newcommand{\charles}[1]{{\color{blue}\textbf{YC: }#1}\normalfont}

    \newcommand{\todo}[1]{{\color{red}#1}}

\else

    \newcommand{\hubert}[1]{{#1}}
    \newcommand{\charles}[1]{{#1}}

    \newcommand{\todo}[1]{{#1}}
    
\fi

\def\paperID{1047} %
\def\confName{CVPR}
\def\confYear{2024}

\title{{V}irtual {P}ets: Animatable Animal Generation in 3D Scenes }

\author{
Yen-Chi Cheng$^{1,3}$
\hspace*{1em}
Chieh Hubert Lin$^{2}$
\hspace*{1em}
Chaoyang Wang$^{3}$
\hspace*{1em}
Yash Kant$^{3,4}$
\\
\hspace*{1em}
Sergey Tulyakov$^{3}$
\hspace*{1em}
Alexander Schwing$^{1}$
\hspace*{1em}
Liangyan Gui$^{1}$
\hspace*{1em}
Hsin-Ying Lee$^{3}$
\\
$^{1}$ UIUC
\hspace*{1em}
$^{2}$ UC Merced
\hspace*{1em}
$^{3}$Snap Research
\hspace*{1em}
$^{4}$University of Toronto
\\
\small \url{https://yccyenchicheng.github.io/VirtualPets/}
}

\begin{document}
\twocolumn[{%
\maketitle
\vspace{-1.0em}
\renewcommand\twocolumn[1][]{#1}%
     \includegraphics[width=\linewidth,trim={0 8cm 0 6cm},clip]{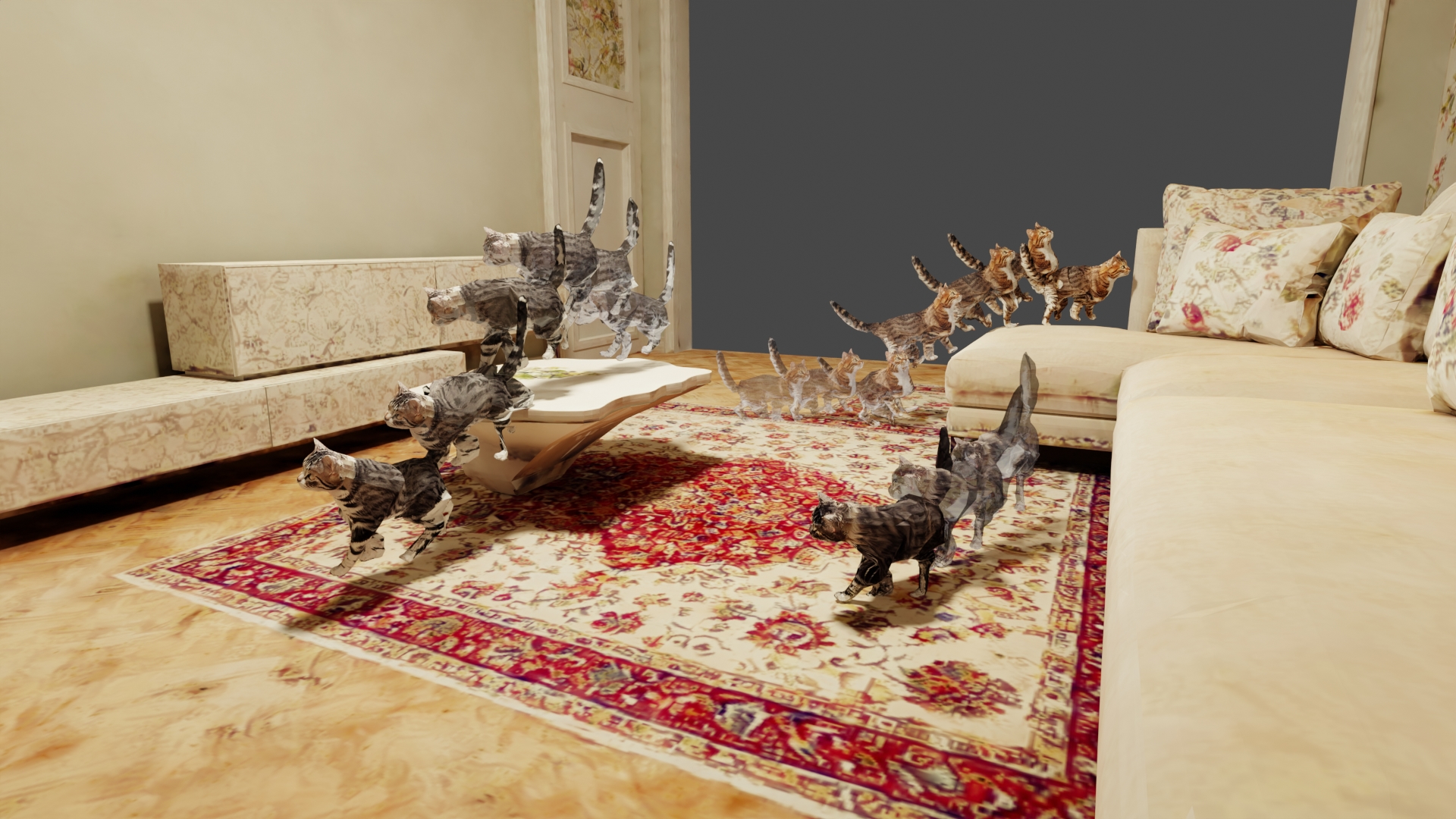}
    \vspace{-2em}
    \captionof{figure}{
        \textbf{Virtual Pets.}
        Given a 3D scene, we can generate diverse 3D animal motion sequences that are environment-aware.
    }
    \label{fig:teaser}
    \vspace{.8em}
}]

\begin{abstract}
Toward unlocking the potential of generative models in immersive 4D experiences, we introduce Virtual Pet, a novel pipeline to model realistic and diverse motions for target animal species within a 3D environment.
To circumvent the limited availability of 3D motion data aligned with environmental geometry, we leverage monocular internet videos and extract deformable NeRF representations for the foreground and static NeRF representations for the background. For this, we develop a reconstruction strategy, encompassing species-level shared template learning and per-video fine-tuning. 
Utilizing the reconstructed data, we then train a conditional 3D motion model to learn the trajectory and articulation of foreground animals in the context of 3D backgrounds.
We showcase the efficacy of our pipeline with comprehensive qualitative and quantitative evaluations using cat videos.
We also demonstrate versatility across unseen cats and indoor environments, producing temporally coherent 4D outputs for enriched virtual experiences.
\end{abstract}

\setlength{\abovedisplayskip}{1.7mm}
\setlength{\belowdisplayskip}{1.7mm}
\setlength{\abovedisplayshortskip}{0pt}
\setlength{\belowdisplayshortskip}{0pt}

\vspace{-2mm}
\section{Introduction}
\vspace{-2mm}
\seclabel{intro}
Recent advances in 3D modeling have noticeably improved the quality of modeling the shape~\cite{Get3D,SDFusion,3dilg,autosdf}, appearance~\cite{text2tex,TEXTure, DreamFusion,Magic3D,Magic123,ProlificDreamer,Fantasia3D}, and dynamics~\cite{singer2023text,li2022neural,park2021nerfies,park2021hypernerf,yang2023reconstructing} for both objects and scenes. 
While these advances are pivotal, there also remains a noticeable gap in %
the vivacity and interactivity inherent in 3D modeling, crucial for a fully immersive experience.
Constructing virtual characters and imbuing them with dynamic, lifelike motions based on their environmental context is pivotal in various domains, including movie production, AR/VR development, and game design.
Currently, these creative workflows demand extensive efforts from skilled artists and 3D designers who meticulously craft each object and manually plan motions. This manual process renders resulting experiences costly, difficult to scale, and time-consuming. Moreover, the designed motions often encounter challenges in seamless transferability to new environments.
In response to these challenges, we propose the development of an environment-aware generative model for 3D motion in a data-driven manner. 
In this work, we specifically focus on modeling \textbf{Virtual Pets}: synthesizing realistic 3D motion for an animal species within diverse 3D scenes.
\comment{
We have seen great advances in 3D shape and appearance modeling, 3D scene reconstruction, and 3D dynamics of deformable objects. Generating or reconstructing faithful 3D representation is crucial for users experience. However, providing controllability over the subject in the scene can provide a much more immersive experience, which is less explored in the current literature.
Controlling a virtual character in a 3D scene is essential in AR/VR, movie industry, and game development. However, building such an animatable character and the world in which it lives is not easy. 
Artists often need to use specialized equipment such as Motion Capture or multiview sensor to capture the targeted motions. To build the synthetic environment it lives in, artistes need to perform complicated 3D modeling and post-processing to construct the world which the motions are compatible. This is expensive, hard to scale, not transferable. However, we have abundant monocular videos captured by amateurs or photographers. Is it possible to extract the motion and the corresponding environment from these videos? In this work, we are mainly focusing on learning environment-aware 3D motion from monocular videos, and then control the subjects we are interested (Cat) in a 3D scene.
}

\vspace{-.5em} %
Developing an environment-aware generative model for 3D motion is challenging due to the scarcity of 3D motion annotations paired with corresponding and detailed environment geometry.
Manual collection of such data with modern motion capture equipment is impractical for individuals due to the associated time and cost.
Additionally, transferring the motions between subjects is difficult, due to  anatomic variations of  subjects and  distinct environment layouts during the capture. 
We, therefore, seek to distill the environment-conditioned motion information from more accessible monocular videos on the Internet, which encompass diverse activities in various environments.
Nevertheless, the problem remains difficult for three primary reasons: (a) traditional structure from motion~\cite{colmap_sfm,colmap_mvs} cannot handle deformable objects, (b) the motions lack transferability without specific remedies, and (c) the motion needs to respect both the anatomy of the subject and the layout of the environment without compromising diversity.

To address these challenges, our approach involves leveraging a collection of videos featuring the target animal species. Our objective is to jointly reconstruct the articulated 3D shape of all foreground objects and to train an environment-aware generative model to produce realistic 3D motions within arbitrary 3D scenes. 
First, we extract 3D shape, motion, and affordance from a collection of videos.
We adopt a deformable NeRF reconstruction method~\cite{yang2023reconstructing} to learn a species-level shared template for the foreground object on all videos. Subsequently, we fine-tune the instance-level deformable NeRF on each video, while the background is jointly reconstructed with a static NeRF, making sure the articulated motion is compatible with the background.
With the reconstructed deformable foregrounds and static backgrounds, we train a conditional 3D motion generation framework to model the trajectory and articulation of the foreground object.
During inference, the model can generate conditional 3D motion sequences given a foreground object and background information, rendering them into coherent videos. 
For textureless foreground or background, we adopt an off-the-shelf texturing method~\cite{text2tex,scenetex} to produce a realistic and diverse appearance.
In this work, we select \textit{cats} as our target species due to their rich deformation in structure and appearance, diverse affordance, and the availability of sufficient video data.

We conduct extensive experiments on a collection of cat videos to demonstrate the effectiveness and efficacy of the proposed pipeline.
We show that the proposed method performs favorably against the baseline methods, both quantitatively and qualitatively. 
Furthermore, we demonstrate our trained model generalizes
to unseen cats and indoor room environments, generating temporally coherent 4D outputs~\ref{fig:teaser}.

\section{Related Work}
\seclabel{related}

\noindent\textbf{Dynamic reconstruction from videos.}
The emergence of Neural Radiance Fields (NeRF)~\cite{mildenhall2020nerf} enabled promising progress in 3D reconstruction from multi-view images.
Compared to images, reconstructing dynamics from videos poses significant challenges in handling time-varying dynamic content. 
Methods based on dynamic radiance fields either require multiple synchronized input videos~\cite{bansal20204d,li2022neural,wang2022fourier,zhang2021editable}, or fail to handle complex dynamic scenarios~\cite{park2021nerfies,park2021hypernerf,xian2021space,li2021neural}.
A stream of work~\cite{yang2023reconstructing,wu2023dove,yang2021lasr,yang2021viser,yang2022banmo} focuses on reconstructing articulated objects from videos, as the prior shape knowledge is known and videos provide more signals to disentangle motion and morphology.
In this work, we adopt RAC~\cite{yang2023reconstructing} as our backbone reconstruction method. 

\noindent\textbf{4D Generation.}
Compared to 2D and 3D data, 4D data is extremely difficult to collect, making it difficult to learn a 4D generative model in a data-driven manner, let alone the environment-aware conditioning that we are interested in.
Recent successes in 3D generation~\cite{DreamFusion,Magic3D,HiFA,Magic123,ProlificDreamer,Fantasia3D} distill knowledge from pretrained large-scale 2D diffusion models. 
Therefore, there are some recent efforts that follow the successes and 
extend the techniques to 4D~\cite{singer2023text,jiang2023consistent4d}. 
However, the quality is not satisfactory and it is difficult to enable additional control and manipulation. 
To bypass the current limitation and focus on the early exploration of the environment-aware 4D generation field, we target a category-specific setting and leverage the state-of-the-art animatable object reconstruction method~\cite{yang2023reconstructing} to provide training data necessary for the proposed pipeline. 

\noindent\textbf{Affordance Prediction.}
Affordance refers to the potential actions or interactions that an object or environment offers.
Early efforts in 2D affordance prediction from images formulate the task as a categorical classification problem~\cite{cai2016understanding,nagarajan2020ego,hermans2011affordance} or a heatmap regression problem~\cite{fang2018demo2vec,huang2018predicting,liu2022joint,nagarajan2019grounded}.
3D affordance prediction is employed to understand human-scene interactions by predicting 3D locations and poses based on 2D images~\cite{fouhey2012people,gupta20113d,li2019putting,zhang2020generating,ye2023affordance} or 3D scenes~\cite{zhao2022compositional,hassan2019resolving,huang2022capturing,zhang2020place}.
In this work, given a 3D scene, we not only want to infer the potential 3D affordance, but also model the plausible movements in a generative manner. 

\noindent\textbf{Motion Prediction.}
Motion prediction has been studied, especially for human motion, using various backbone architectures from early Markov models~\cite{lehrmann2014efficient}, recurrent models~\cite{fragkiadaki2015recurrent,jain2016structural,martinez2017human}, generative adversarial networks~\cite{barsoum2018hp,lee2019dancing}, to the recent diffusion models~\cite{tevet2022human}.
The prediction can be conditioned on action class~\cite{guo2020action2motion,petrovich2021action}, audio~\cite{lee2019dancing,li2021ai}, and text~\cite{tevet2022human,ahuja2019language2pose}. 
However, most existing motion prediction methods are based on skeletons, disregarding surface geometry and appearance.

\begin{figure*}[t!]
    \centering
    \includegraphics[width=1.0\linewidth]{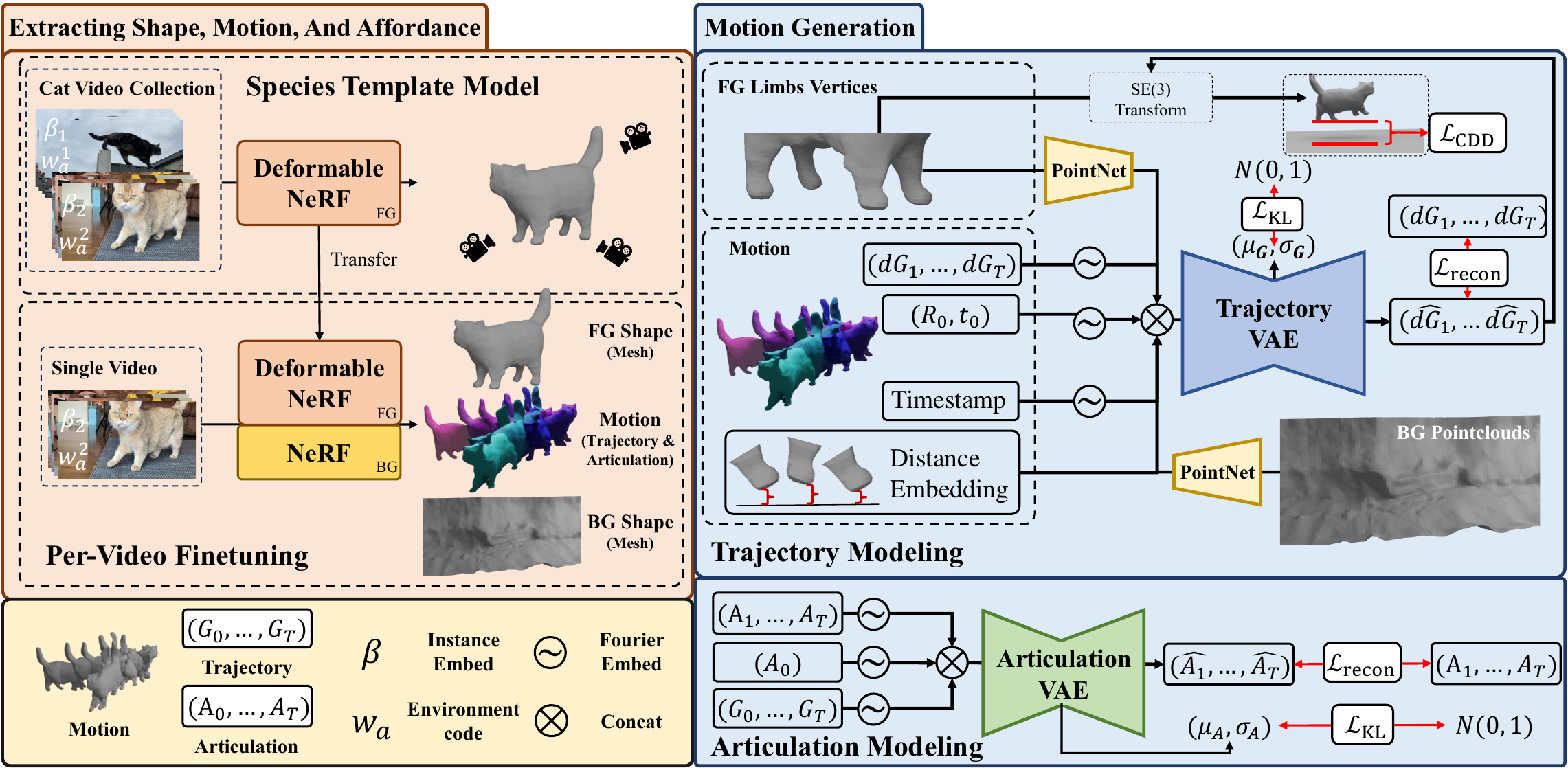}
    
    \vspace{-.5em}
    \caption{
        \textbf{The proposed framework of Virtual Pets.} 
        To extract 3D shapes and motions from monocular videos: 
         we first learn a \textit{Species Articulated Template Model} with an articulated NeRF~\cite{yang2023reconstructing} using a collection of cat videos. 
        We then perform \textit{Per-Video Fine-tuning}. For each video, we further reconstruct the background with a static NeRF~\cite{mildenhall2020nerf}. The articulated NeRF trained in the species-level stage is loaded and fine-tuned in this stage to make sure the motions, which are Trajectory and Articulation, respect the reconstructed background shape. After that, we train an environment-aware 3D motion generator with a \textit{Trajectory VAE} and an \textit{Articulation VAE}. It generates 3D motions based on vertices of the foreground limbs, distance from foreground to background, and pointclouds sampled from the background.
        }

    \vspace \figmargin
    \label{fig:2.overview.p2}
\end{figure*}

\section{Method}
In the following, we first describe how to extract shape, motion, and affordance from an unstructured monocular video collection (see Sec.~\ref{sec:3.rac}). We then discuss our environment-aware generative model for 3D motion (see Sec.~\ref{sec:4.approach}).

\subsection{
Extracting Shape, Motion, And Affordance Using Articulated 3D Reconstruction
}
\label{sec:3.rac}
\noindent\textbf{Overview.}
To develop an environment-aware generative model for 3D motion, we extract the relevant representations from an unstructured monocular video collection in two stages, as illustrated in the left column of Fig.~\ref{fig:2.overview.p2}.
In the first stage (see Sec.~\ref{sec:prelim-template-sample} for more), we focus on learning a species-level articulated template shape.
This template shape serves as a foundational element, allowing subsequent alignment of motions extracted per video in the following stage. This alignment process significantly aids in the training of the generative model. 
Then, in the second stage (see Sec.~\ref{sec:prelim-finetuning} for more), we transfer and fine-tune the template shape on a per-video basis to obtain a more  accurate object shape $H_\text{fg}$, trajectory $\{G_t\}$, and articulations $\{A_t\}$, where $t$ denotes the time in the given video. 
Alongside fine-tuning the template shape, we simultaneously learn the background scene shape $H_\text{bg}$ by learning a static NeRF model.
The affordances of the objects in the environment can then be represented as interactions between the salient object shape $H_\text{fg}$ and the background scene shape $H_\text{bg}$.
This two-stage procedure enables to extract paired 3D motion and environment models from monocular videos.

\comment{
\charles{
Given a category of monocular videos, our goal is to reconstruct: (1) foreground object's shape $H_\text{fg}$, (2) background scene's shape $H_\text{bg}$, (3) foreground object's motion. It contains three steps: (1) reconstructing template shape for foreground object from a collection of 2D videos. (2) fine-tuned on each video to reconstruct FG shape, BG shape for all timesteps. (3) Extract FG shape, BG shape, and FG motion. Please refer to Fig.~\ref{fig:2.overview.p1} for a summary of this stage.
}
}

\subsubsection{Species Articulated Template Shape}
\label{sec:prelim-template-sample}

Given videos of one animal species (in our case, cats), we learn a shared template shape. 
We represent the shared template shape via $(c^t, d, G_t, A_t)$, encompassing the color $c^t$, the signed distance function (SDF) value $d$, the object pose $G_t$, and the articulation $A_t$ of the shape. 
The object pose $G_t = (R_t, s_t) \in SE(3)$ describes the object pose in the camera space, where rotation $R_t$ and translation $s_t$ transform the object from its rest pose in canonical space to the pose in camera space. We use a sequence of object poses $\bfG = [G_0, G_1, \ldots, G_T]$ to characterize the global movements of the foreground object in the environment. 
Articulation $A_t \in \mathbb{R}^{3 \times B}$ expresses the local changes of $B$ skeleton joints that deform the template model at every time step.
To encapsulate all the global-local time-dependent spatial dynamics of the shape, we represent motion as ``$M_t = (G_t, A_t)$''.

To model the template shape, we use an implicit function $F_{\theta}$, concretely, an articulated NeRF~\cite{yang2023reconstructing}. 
Given any 3d point $\bfx$ in space $\mathbb{R}^3$, for any time $t$, for any instance embedding $\beta \in \mathbb{R}^{128}$, and for any scene-dependent environment feature $w_a$, the implicit function returns the template shape $(c^t, d, G_t, A_t)$:
\begin{equation}
    (c^t, d, G_t, A_t) = F_{\theta}(\bfx, \beta, w_a) \, .
\end{equation}
Both $\beta$ and $w_a$ are trainable parameters to learn the video-specific appearance such as the identity of the subject and the lighting condition of the environment.

\comment{
\charles{
Given a category of monocular videos, we reconstruct a template of shape for foreground objects $H_\text{fg}$ of different instances in each video. We adopt a deformable NeRF~\cite{mildenhall2020nerf,yang2023reconstructing}. The goal is to learn a deformable radiance field $F_{\theta}: (\bfx, \beta, w_a) \rightarrow (c^t, d, G_t, A_t)$, which obtain the color $c^t$ and SDF $d$ at $\bfx$ canonical space at time $t$, where $\beta$ is a instance embedding, and $w_a$ is a environment code in different video. And the motion $M_t = (G_t, A_t)$, where $G_t = (R_t, s_t) \in SE(3)$ transform $H_\text{fg}$ from canonical space to camera space. $G_t$ describes the trajectory (global motion) of the foreground object. $A_t \in SE(3)$ \todo{dimension of $A_t$?} articulation (local motion) at time $t$ to deform the shape from canonical space to camera space. We discuss three major components to describe a deformable object: shape and appearance, skeleton, and skinning field. \\
}
}

$F_{\theta}$ is implemented with a combination of a skeleton, skinning weights $W$ driven by the skeleton, and multiple Multi-Layer Perceptrons (MLPs) representing the shape and appearance deformed by the skinning weights.
In the following, we describe the three components of the implicit function $F_{\theta}$.

\noindent\textbf{Shape and appearance.} 
We encode canonical coordinate $\bfx \in \mathbb{R}^3$, viewing direction $\bfv^t$, instance embedding $\beta \in \mathbb{R}^{128}$, and environment embedding $w_a$ into color $c^t$ and SDF value $d$ via an MLP, \ie, 
\begin{align}
    c^{t} &= \mathrm{MLP}_{\mathrm{c}}(\bfx, \bfv^t, w_a)\,, \label{eq:fg_nerf_c} \\
    d &= \mathrm{MLP}_{\mathrm{SDF}}(\bfx)\,,\,\,\, \sigma = \Gamma(d)\,, \label{eq:fg_nerf_d}
\end{align}
where $\Gamma$ is the CDF of a Laplacian distribution following the design of VolSDF~\cite{yariv2021volume}.%

\comment{
\noindent\textbf{Shape and Appearance.} In the canonical space, a 3D point $\bfx \in \mathbb{R}^3$ will be encoded to three features: SDF, color, and canonical feature. To model the difference between instances in a category, an instance embedding $\beta$ is introduced to encode shape and appearance. 
\begin{align}
    c^{t} &= \mathrm{MLP}_{\mathrm{c}}(\bfx, \v^t, w_a)\, \label{eq:fg_nerf_c}, \\
    d &= \mathrm{MLP}_{\mathrm{SDF}}(\bfx)\,, \label{eq:fg_nerf_d} \\
    \psi &= \mathrm{MLP}_{\psi}(\bfx)\,. \nonumber
\end{align}
where $w_a$ is an environment code to capture the conditions such as illumination and shadows, and $CDF$ is \todo{condtion for which CDF can be use} used to obtain the $sigma$ for volume rendering. We choose Laplacian's CDF following the design VolSDF~\todo{cite}.\\
}

\noindent\textbf{Articulation.} 
The articulation $A_t$ is conditioned on instance embedding $\beta$, and is learned by modeling the joints $J_{\alpha} \in \mathbb{R}^{3 \times B}$ and skinning weights $W_{\beta}$ for each video. 
The intuition of articulated 3D reconstruction is to derive a skeleton along with skinning weights, so that the sparse set of transition vectors (\ie, articulation) applied on the skeleton joints can be converted into a dense motion field specifying the deformation of any point in the space over time.
Following RAC~\cite{yang2023reconstructing}, we define the skeleton topology of the template shape as a collection of $(B+1)$ bones connected by $B$ joints.
The dynamics of the bones are determined by two learnable per-video embeddings: (1) the instance embedding $\beta$ and (2) the location of the bone.
We model the location of the joints via  
\begin{align}
    J = \mathrm{MLP}_{J}(\beta) \in \mathbb{R}^{3 \times B} \, ,
\end{align}
encapsulating the movements of the bones.
The skinning weights $W \in \mathbb{R}^{B+1}$ then specify the deformation factors of a canonical coordinate depending on the changes of all bones via  
\begin{align}
    W = \sigma_{\mathrm{softmax}} ( d_{\sigma}(\bfx, \beta) + \mathrm{MLP}_{W} (\bfx, \beta) \, .
\end{align}
Here, $d_{\sigma}(\bfx, \beta, \theta)$ is the Mahalanobis distance between the 3D point $\bfx$ and the bones, following the design of BANMo~\cite{yang2022banmo}.
Finally, we obtain the articulation $A_t$ with
\begin{equation}
    A_t = \mathrm{MLP}_{A}(\beta, t) \in \mathbb{R}^{3 \times B}\, ,
    \label{eq:articulation}
\end{equation}
which  represents the deformations of each bone.
The articulation at each time step can be converted into a dense warp field $D$ using dual quaternion blend skinning (QBS)~\cite{yang2023reconstructing,kavan2007skinning}.
We denote this time-dependent dense warp field as
\begin{align}
    D_t = \text{QBS}(J, W, A_t, F_\theta)\,.
    \label{eq:warping_field}
\end{align}

\noindent\textbf{Object Pose.}
The object pose $G_t$ is computed by querying a time-conditioned MLP. It computes $(R_t, s_t)$ to transform from an object's rest pose in canonical space to the observed object pose in the camera space via 
\begin{align}
    G_t = \mathrm{MLP}_\text{pose} (t)\,.
    \label{eq:obj_pose}
\end{align}

\noindent\textbf{Rendering and reconstructing.} 
We can now perform volume rendering to render the images for NeRF training given $(G_t, A_t)$.
Since the shape and appearance is modeled in canonical space, we first use $G_t$ to transform the observed points to camera space. At time $t$, let $x_c^t \in \mathbb{R}^2$ refer to an observed pixel in camera space. $\bfx_c^{t} \in \mathbb{R}^3$ is a 3D point sampled along the ray starting at the camera center and passing through $x_c^t$. Using the object pose $G_t^{-1}$, we transform $\bfx_c^{t}$  back to a point $\bfx$. We then use the articulation $A_t$ to compute the dense warp field $D_t$ using Eq.~\eqref{eq:warping_field} to warp $\bfx$ and query the MLP to obtain $(\sigma, c^t)$ by querying $D_t(\bfx)$ using $\mathrm{MLP}_{\mathrm{c}}$ and $\mathrm{MLP}_{\mathrm{SDF}}$ with Eq.~\eqref{eq:fg_nerf_c} and Eq.~\eqref{eq:fg_nerf_d}. We can then perform volume rendering with $(\sigma, c^t)$ following a conventional NeRF.

\subsubsection{Per-video finetuning}
\label{sec:prelim-finetuning}

Given a monocular video, we aim to reconstruct both the foreground mesh $H_\text{fg}$ and the background mesh $H_\text{bg}$. To make sure the reconstructed mesh $H_\text{fg}$ is compatible with mesh $H_\text{bg}$,
we optimize the Deformable NeRF $\mathrm{F}_{\theta}$, initialized with the optimized $\mathrm{F}_{\theta}$ in the articulated template learning stage, and background NeRF $\mathrm{F}_{\phi}$ simultaneously. 
Specifically, given a 3D point $\bfx$ and viewing direction $v$, we optimize foreground and background NeRFs for each video to reconstruct the full scene: 
\begin{align}
\begin{split}
    (d, c^t)_{\text{bg}} = \mathrm{F}_{\phi}(\bfx, v) \, , \hspace{4mm}
    (d, c^t)_{\text{fg}} = \mathrm{F}_{\theta}(\bfx, \beta, w_a)\, ,
\end{split}
\end{align}
where $\sigma$ is the density and $c^t$ is the color at time $t$ for 3D location $\bfx$. The final rendering  after aggregation is determined by the depths estimated with $\sigma$.

After the species-level template shape learning and the per-video foreground and background joint optimization, we can extract the 3D motion $M_t = (G_t, A_t)$, foreground mesh $H_\text{fg}$, and background mesh $H_\text{bg}$ with $F_{\theta}$ and $F_{\phi}$ given each video.
The mesh for foreground and background is obtained by querying $\mathrm{F}_{\theta}$ and $\mathrm{F}_{\phi}$ to get the SDF in the canonical space, and by using a marching cube~\cite{lorensen1998marching} to extract the shape at a level-set equalling $0$. Pose $G_t$ is obtained via Eq.~\eqref{eq:obj_pose}, and articulation $A_t$ is computed with Eq.~\eqref{eq:articulation}.

\begin{figure*}[t!]
    \centering
        \includegraphics[width=\linewidth]{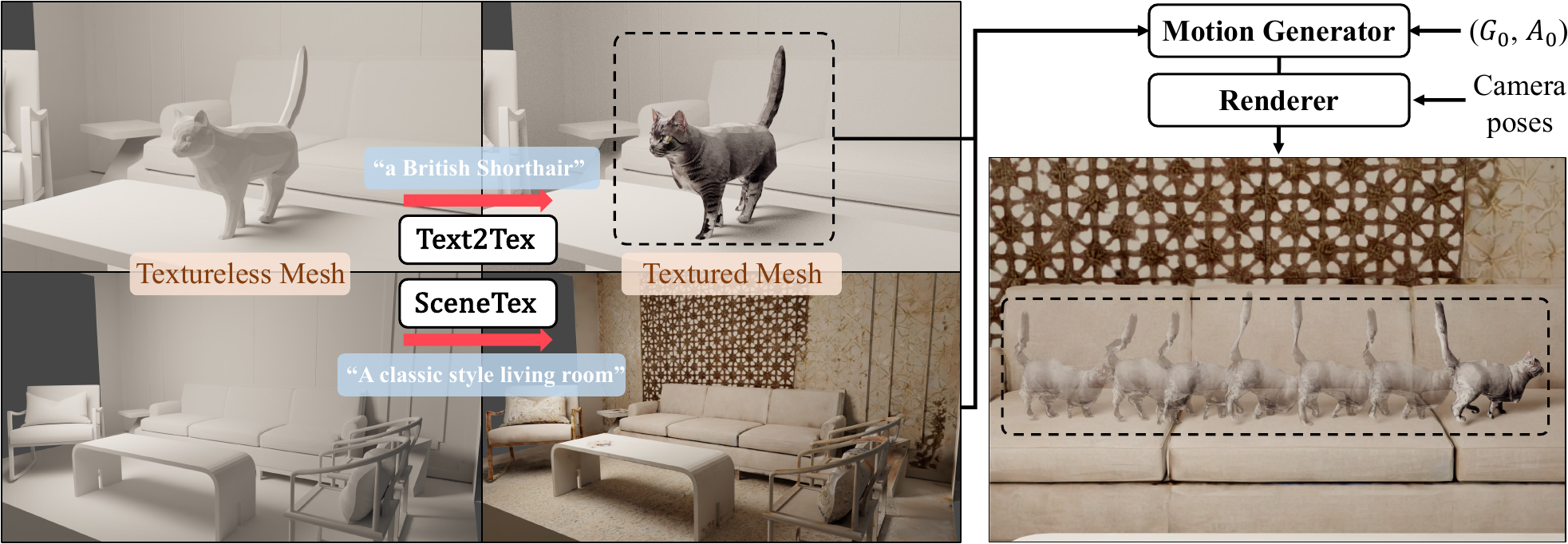}
        \vspace{-2em}
    \caption{
        \textbf{Texturing and Rendering.}
        At  inference time, given textureless foreground and background meshes, we first adopt Text2Tex~\cite{text2tex} and SceneTex~\cite{scenetex} to texture the meshes. 
        Meanwhile, we generate the motion sequence using the trained trajectory VAE and articulation VAE. 
        We then obtain the final predicted foreground mesh after deformation and transformation. 
        Finally, the 3D motion sequences and the 3D scene and be rendered to videos.
        }
    \vspace \figmargin
    \vspace{-3mm}
    \label{fig:2.overview.p3}
\end{figure*}

\subsection{Approach: Motion Generation}
\label{sec:4.approach}

Given the extracted foreground mesh $H_\text{fg}$, the background mesh $H_\text{bg}$, and the motion $M_t$, our goal is to learn a conditional 3D motion generative model that can synthesize a 3D motion sequence of a cat in an indoor scene. 
We formulate the 4D generation in two stages. 
In the first stage (Sec.~\ref{ssec:4.1.vae}), we generate the 3D motion trajectory given a cat (\ie, foreground mesh) and a scene (\ie, background mesh) using a `Trajectory VAE'. %
Then, given the trajectory, we predict the corresponding articulation using an `Articulation VAE' . 
The right column of Fig.~\ref{fig:2.overview.p2} provides an overview of the proposed `Trajectory VAE' and the `Articulation VAE' for motion generation. 
In the second stage, as the foreground and background meshes are textureless, we adopt an off-the-shelf text-to-image diffusion model and score distillation sampling to perform texturing. 
Finally, we use the motion generated by our motion model to transform and deform the textured cat mesh for each timestep and render the 4D output into a video (Sec.~\ref{ssec:4.2.render}). Fig.~\ref{fig:2.overview.p3} shows the overall inference pipeline given foreground and background meshes, including texturing and rendering.

\subsubsection{Learning Environment-Aware 3D Motion}
\label{ssec:4.1.vae}

Given a foreground mesh $H_\text{fg}$ and a background mesh $H_\text{bg}$, we leverage a conditional VAE to generation the 3D motion $(\bfG, \bfA) = ((G_0, A_0), (G_2, A_2), \ldots, (G_T, A_T) ) $. We generate the motion with two conditonal VAEs -- the Trajectory VAE ($\mathrm{VAE}_{\bfG}$) and the  Articulation VAE ($\mathrm{VAE}_{\bfA}$). $\mathrm{VAE}_{\bfG}$ first generates an environment-aware trajectory $\bfG$. Subsequently, the Articulation VAE ($\mathrm{VAE}_{\bfA}$) takes the trajectory as input to generate the corresponding articulation $\bfA$. \\
\noindent\textbf{Trajectory VAE.} Given the starting pose $G_0 = (R_0, s_0)$, we first generate the global trajectory $d\mathbf{G} = (d G_1, d G_2, \ldots, d G_T) $, where $d G_t$ is the relative rotation and translation at time $t$. Note, $R_0, s_0$ serve as the reference pose. %
For environment-awareness we additionally condition on (1) the vertices of the foreground limbs $\bfP_\text{limb} \in \mathbb{R}^{N_\text{fg} \times 3}$ ($N_\text{fg}$ is number of vertices to sample from $H_\text{fg}$), (2) the distance $D_\text{fg} \in \mathbb{R}$ of the center of the foreground mesh to the background mesh $H_\text{bg}$, and (3) a pointcloud $\bfP_\text{bg} \in \mathbb{R}^{N_\text{bg} \times 3}$ sampled from $H_\text{bg}$. We use the PointNet~\cite{qi2017pointnet} encoder to encode $\bfP_\text{limb}$ and $\bfP_\text{bg}$. 

Formally, our Trajectory VAE encoder first computes embeddings of the pointcloud data via
\begin{align}
    \bfz_\text{bg} &= \mathrm{PointNet} (\bfP_\text{bg})\,, \,\,\, \bfz_\text{limb} = \mathrm{PointNet} (\bfP_\text{limb}).
\end{align}
We also embed the starting pose $G_0$ and the trajectory sequence extracted from video data via MLPs, i.e.,
\begin{align}
\begin{split}
\bfz_{G_0} &= \mathrm{MLP}_{G} ( G_0 ),\\
\bfz_{d \bfG} &= \mathrm{MLP}_{ dG } ( d G_1, d G_2, \ldots, d G_T ).
\end{split}
\end{align}
Here, $\mathrm{MLP}_{ dG }$ is an encoder implemented via an MLP to encode the features of the sequence $(dG_t)_{t=1}^{T}, dG_t \in SE(3)$.
A latent variable encoding the trajectory is sampled from the normal distribution with mean and standard deviation obtained from an encoder  $\mathbf{E}_{\bfG}$ implemented with a ConvNet and an MLP:
\begin{align}
 \mu_{\bfG}, \sigma_{\bfG} = \mathbf{E}_{\bfG}(\bfz_{ d \bfG }, \bfz_{G_0}, \bfz_\text{limb}, \bfz_\text{bg}, D_\text{fg}, \mathbf{\tau})\, ,
\end{align}
where $\bfz_{\bfG} \sim N(\mu_{\bfG}, \sigma_{\bfG})$. We use the decoder to generate a trajectory:
\begin{align}
    \hat{d \bfG} &= \mathbf{D}_{\bfG}(\bfz_{\bfG}, \bfz_{G_0}, \bfz_{fg}, \bfz_{bg}, \tau),
\end{align}
where $\tau = (\tau_{0}, \tau_{1}, \ldots, \tau_{T}) \in \mathbb{R}^{T \times 128}$ is a sequence of time-embeddings to enhance the temporal awareness. Note, we use the Fourier embeddings of $G_0$, $d \bfG$, and $\tau$ for better reconstruction results.

\begin{figure*}[t]
    \centering

    \includegraphics[width=1.0\linewidth]{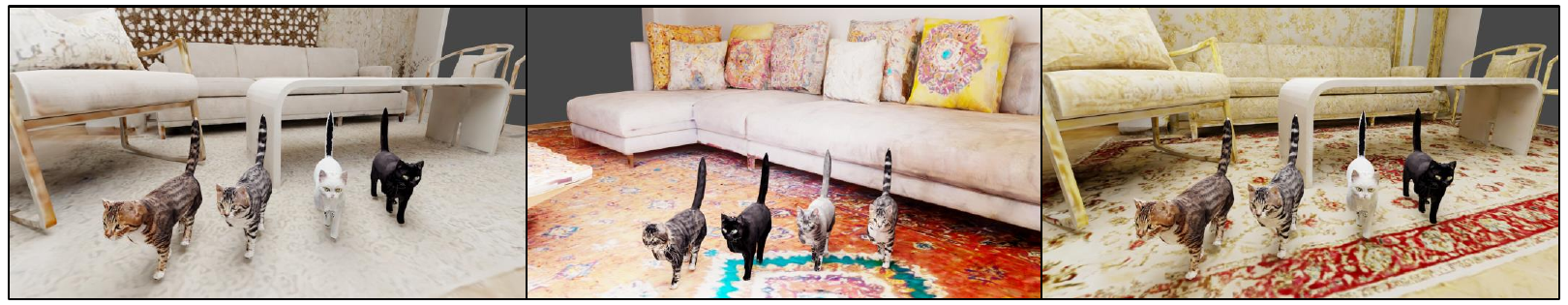} %
    
    \vspace{-2mm}
    \caption{
    \textbf{Diverse textures.}
    We adopt Text2Tex~\cite{text2tex} and SceneTex~\cite{scenetex} to perform diverse texturing to both foreground objects and background scenes.
    }
    \vspace{-3mm}
    \vspace \figmargin
    \label{fig:4.add_texture}
\end{figure*}

\noindent\textbf{Articulation VAE.} Given the trajectory $\bfG$ and the starting articulation $A_0$, $\mathrm{VAE}_{\bfA}$ generates the corresponding  articulation sequence $\bfA$. For each motion sequence, we first encode the trajectory $\bfG$ and $\bfA$ into latent codes
\begin{align}
\begin{split}
    (\bfz_{G_0}, \bfz_{G_1}, \ldots, \bfz_{G_T}) &= \mathrm{MLP}_{ G } ( G_0, G_1, \ldots, G_T ),\, \\
    \bfz_{\bfA} &= \mathbf{E}_{ \bfA } ( A_1, A_2, \ldots, A_T ),\, \\
    \bfz_{A_0} &= \mathrm{MLP}_{ A} (A_0)\, .
\end{split}
\end{align}
Here, $\mathrm{MLP}_{G}$ is an MLP to encode the trajectory $(G_{t})_{t=0}^{T}$.
We also embed the starting articulation $A_0$ with $\mathrm{MLP}_{A}$.
The means and standard deviations are obtained via:
\begin{align}
    \mu_{\bfA}, \sigma_{\bfA} &= \mathbf{E}_{\bfA} \left( (\bfz_{ G_{t} })_{t=0}^{T}, \bfz_{ \bfA }, \bfz_{A_0}, \mathbf{\tau} \right)\,, 
\end{align}
where $\mathbf{E}_{\bfA}$ is an encoder implemented with a ConvNet and MLPs. We then sample a latent variable $\bfz_{\bfA} \sim N(\mu_{\bfA}, \sigma_{\bfA})$ that encodes the articulations from the normal distribution, and decode the articulation:
\begin{align}
\begin{split} 
    \hat{\bfA} = \mathbf{D}_{\bfA}(\bfz_{ \bfA }, \bfz_{A_0}, \bfz_{ \bfG }, \mathbf{\tau})\, . 
\end{split}
\end{align}

\noindent\textbf{Training, Loss, and Inference.}
We train the model with a standard VAE objective: reconstruction loss and KL-divergence loss. Hence, training of $\mathrm{VAE}_{\bfG}$ and $\mathrm{VAE}_{\bfA}$ optimizes the ELBO, i.e.,
\begin{align}
    \mathcal{L}_{\mathrm{VAE}_{\bfG}} &= \mathbb{E}_{q (\bfz_\bfG | \bfG, \bfc_{\bfG})} \log p (\bfG | \bfz_{\bfG}, \bfc_{\bfG}) \nonumber \\
    &- \lambda^{KL}_{\bfG} D_{\mathrm{KL} (q (\bfz_\bfG | \bfG, \bfc_{\bfG}) || p(\bfz_\bfG | \bfc_\bfG) ) }\,, \\
    \mathcal{L}_{\mathrm{VAE}_{\bfA}} &= \mathbb{E}_{q (\bfz_\bfA | \bfA, \bfc_{\bfA})} \log p (\bfA | \bfz_{\bfA}, \bfc_{\bfA}) \nonumber \\
    &- \lambda^{KL}_{\bfA} D_{\mathrm{KL} (q (\bfz_\bfA | \bfA, \bfc_{\bfA}) || p(\bfz_\bfA | \bfc_\bfA) ) }\,.
\end{align}
Here, $\bfc_{\bfG}$ and $\bfc_{\bfA}$ are the conditioned inputs for both VAEs. $\lambda^{KL}_{\bfG}, \lambda^{KL}_{\bfA}$ are the weights for the KL-divergence loss. 

To reduce the floating phenomenon, where foreground objects are unnaturally detached from the ground, 
we add a  `Floating Loss' $\mathcal{L}_\text{CDD}$ while training $\mathrm{VAE}_{\bfG}$.
Given the foreground limb vertices $\bfP_\text{limb}$, we transform it with ground-truth trajectory $\bfG_\text{GT}$ and predicted trajectory $\bfG_\text{pred}$:
\begin{align}
\begin{split}
    \bfG_\text{GT} \bfP_\text{limb} &= [G^\text{GT}_0 \bfP_\text{limb}, G^\text{GT}_1 \bfP_\text{limb}, \ldots, G^\text{GT}_T \bfP_\text{limb}]\,, \\
    \bfG_\text{pred} \bfP_\text{limb} &= [G^\text{pred}_0 \bfP_\text{limb}, G^\text{pred}_1 \bfP_\text{limb}, \ldots, G^\text{pred}_T \bfP_\text{limb}]\,.
\end{split}
\end{align}
We then compute the Chamfer Distance between (1) the transformed vertices at each timestep and (2) background pointcloud $\bfP_\text{bg}$. Formally, 
\begin{align}
    \mathcal{L}_\text{CDD} = \frac{1}{T} \sum_{t=0}^{T} \, \Big\lVert \, & \mathrm{Chamfer} (G^\text{GT}_t \bfP_\text{limb}, \bfP_\text{bg})\, \nonumber \\
     & - \mathrm{Chamfer} (G^\text{pred}_t \bfP_\text{limb}, \bfP_\text{bg}) \Big\rVert_1 \, .
\end{align}
Notice that we compute the distance between the ground-truth and predicted Chamfer Distance instead of directly using $(G^\text{pred}_t \bfP_\text{limb} , \bfP_\text{bg})$ in the objective. 
Solely minimizing the Chamfer Distance between vertices and background will jeopardize motions like jumping. 

In summary, the final objective is: 
\begin{align}
    \mathcal{L} &= \mathcal{L}_{\mathrm{VAE}_{\bfG}} + \lambda_\text{CDD} \mathcal{L}_\text{CDD} + \mathcal{L}_{\mathrm{VAE}_{\bfA}}\,.
\end{align}

\subsubsection{4D Generation and Video  Rendering}
\label{ssec:4.2.render}
Given foreground mesh $H_\text{fg}$ and background mesh $H_\text{bg}$ without textures, 
we first adopt an off-the-shelf 3D texturing method Text2Tex~\cite{text2tex}, to obtain the textured UV-map for both foreground and background meshes.
Then, we generate the motion sequence $(\bfG, \bfA)$.  
Users can specify the starting position and pose for the foreground mesh with $G_0 = (R_0, s_0)$.
We then use $\mathrm{VAE}_{\bfG}$ and $\mathrm{VAE}_{\bfA}$ to generate the motion sequence $(\bfG^\text{gen}, \bfA^\text{gen})$. 
With $\bfA^\text{gen}$, we can adopt Eq.~\eqref{eq:warping_field} to compute the warping field to deform the vertices of $H_\text{fg}$. 
Applying $\bfG^\text{gen}$ afterward can transform $H_\text{fg}$ from canonical space to camera space in different timesteps $H_\text{fg}^{t}$. 
Finally, we combine $[(H_\text{fg}^0, H_\text{bg}), (H_\text{fg}^1, H_\text{bg}), \ldots, (H_\text{fg}^T, H_\text{bg})]\}$ and use a renderer (\eg, PyTorch3D~\cite{ravi2020pytorch3d}, Blender~\cite{blender}) to render the final video.
Fig.~\ref{fig:2.overview.p3} illustrates the whole inference pipeline.

\begin{figure*}[t]
    \centering
    \includegraphics[width=\linewidth]{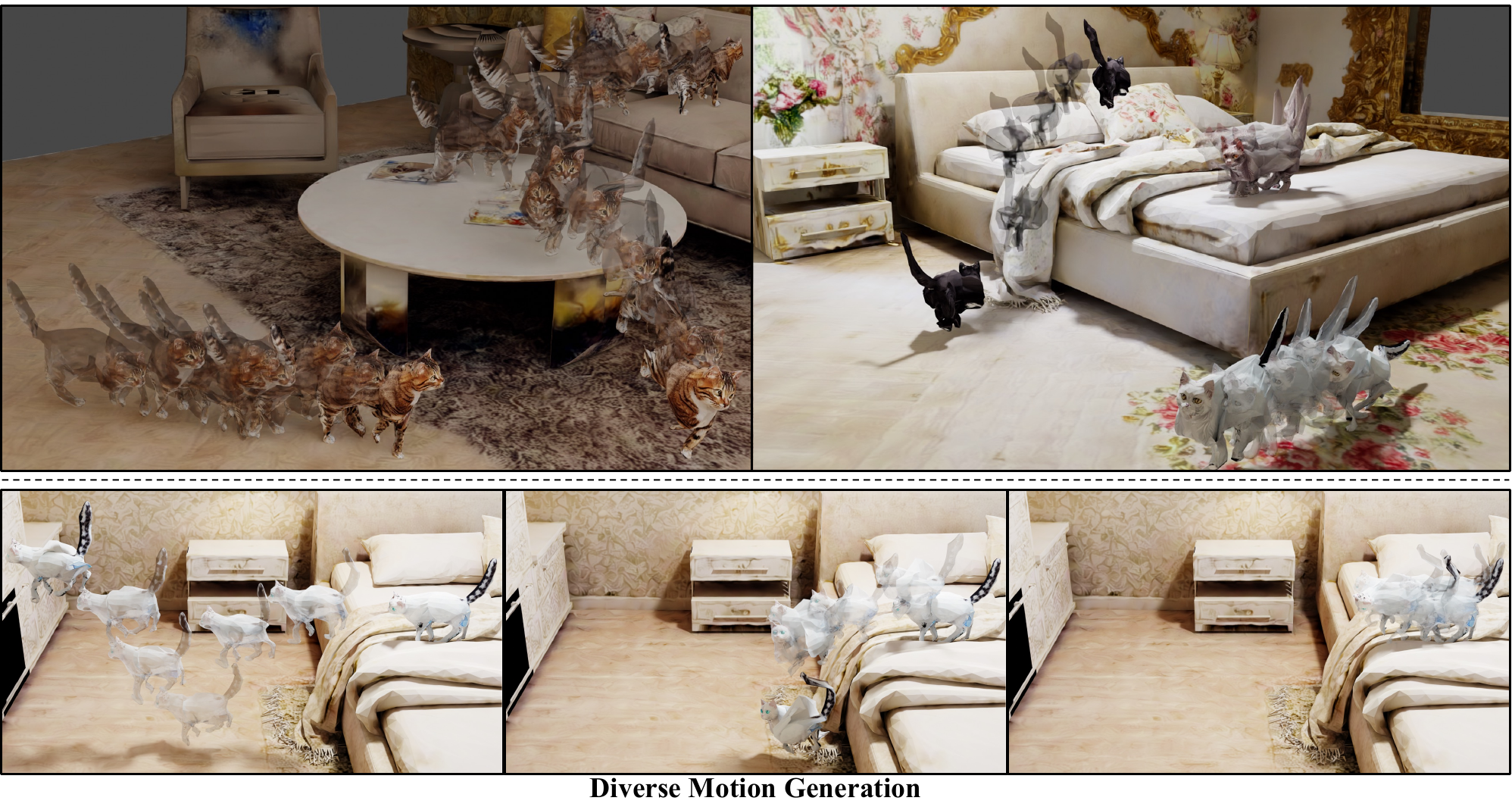}
    
     \vspace{-2mm}
    \caption{
    \textbf{Diverse Environment-aware Motion Generation.} We show the diverse motion generations in different environments. 
    \textit{(Top)} We show 4D generation given different starting poses $G_0$. 
    \textit{(Bottom)} We show diverse motion outputs given the same starting pose in the same scene. The proposed method can generate diverse motions in different environments.
    }
     \vspace{-2mm}
    \label{fig:4.render_ours}
\end{figure*}

\subsection{Implementation Details}
For the stage of learning the species template model, we train for 120 epochs and use a learning rate of $5e^{-4}$ with the AdamW~\cite{loshchilov2017decoupled} optimizer. For the stage of per-video fine-tuning, we train the NeRF using a learning rate of $1e^{-5}$ with the AdamW optimizer for 120 epochs. For training the motion generator, we use a batch size of $16$, a learning rate of $5e^{-4}$, and the Adam~\cite{KingmaAdamICLR} optimizer. The weights for the KL-divergence loss are $\lambda_{\bfG}^{\mathrm{KL}} = 1e^{-2}$, $\lambda_{\bfA}^{\mathrm{KL}} = 1e^{-4}$, respectively, and the weight for the floating loss is $\lambda_\text{CDD}=0.1$ We adopt data augmentation by applying random rotation and translation for the 3D motion, foreground and background shape. For testing the model in unseen environment meshes, we align the mesh to the same orientation and scale as the training background mesh. For texturing the mesh using SceneTex~\cite{scenetex}, we use $40$ update steps for all prompts. We use PyTorch3D~\cite{ravi2020pytorch3d} to render the scene when computing FID and  Blender~\cite{blender} to render the meshes.

\section{Experiments}
\seclabel{experiment}
We conduct qualitative and quantitative experiments to evaluate the efficacy of the proposed pipeline.
We evaluate the quality and diversity of different methods on a collection of cat videos from RAC~\cite{yang2023reconstructing} and $200$ additional collected videos.  
We also render the 4D generation into videos to evaluate its quality.

\noindent\textbf{Data.} For extracting paired 3D motion and environment data, we preprocess each video following RAC~\cite{yang2023reconstructing}. For training the motion generation network, we randomly sample a clip of $T$ frames for each motion sequence extracted from the video. We also translate the background such that its centroid is at the origin $(0, 0, 0)$, and shift the corresponding motion by the same translation. The cat and room meshes for testing are obtained from Turbosquid~\cite{turbosquid} and 3D-Front~\cite{3dfront}. 
We normalize them with the statistics of the scale and center based on $H_\text{fg}$ and $H_\text{bg}$ we obtained from the species-level articulated template shape learning stage.

\begin{figure*}[t]
    \centering
    \includegraphics[width=\linewidth]{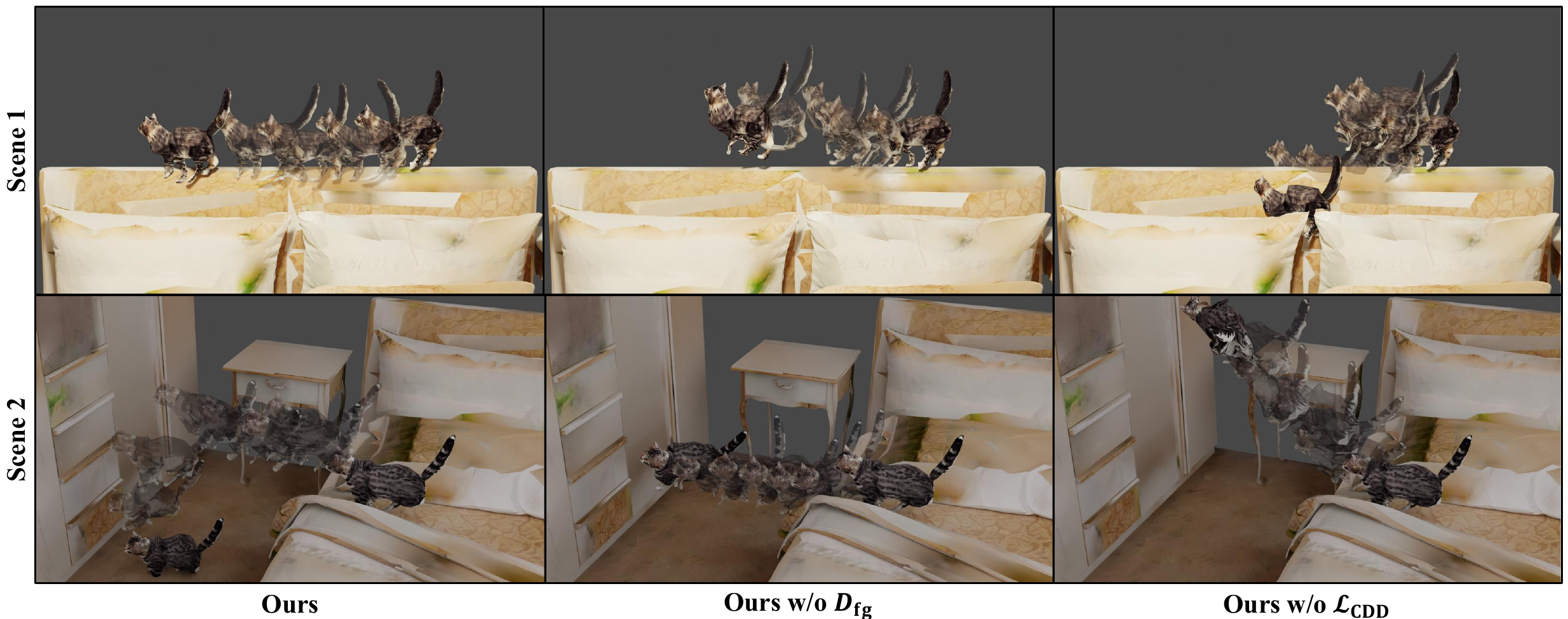}
    \vspace{-1.5em}
    \caption{
    \textbf{Ablation analysis.}
    We ablate the necessity of conditioning on the distance of the foreground mesh center to the background mesh $D_\text{fg}$ and the floating loss $\mathcal{L}_\text{CDD}$. 
    Without $D_\text{fg}$ and $\mathcal{L}_\text{CDD}$, the generated motions cannot faithfully respect the geometry of the 3D scenes and physic constraints. 
    }
    \vspace{-1mm}
    \vspace \figmargin
    \label{fig:4.render_comp}
\end{figure*} 

\subsection{Quantitative Evaluation}
We evaluate the motion generation model in terms of quality and diversity. We also evaluate the quality of 4D rendering results by testing the model on unseen cat meshes and a novel background mesh. \\
\noindent\textbf{Metrics.} For motion generation, we compute the reconstruction error (Recon.) and the floating error (Floating Err.) to measure the quality. We compute the L1 distance of the reconstructed trajectory and the ground-truth trajectory for the reconstruction error. To evaluate the floating error, we use the Chamfer distance between the foreground vertices of limb $\bfP_\text{limb}$ and the background point-clouds $\bfP_\text{bg}$, excluding the jumping sequence. Chamfer distance is computed by finding the nearest-neighbor matches between points in $\bfP_\text{limb}$ and the points in $\bfP_\text{bg}$. For diversity, for each ground-truth trajectory, we generate $N$ motions and compute the $L1$ distance between the ground-truth and the generated trajectories. For rendering quality, we use Fréchet inception distance (FID)~\cite{heusel2017gans} to assess the quality. We compute FID between the rendering of the 4D output generated by our method against the rendering of the ground-truth motion. \\
\noindent\textbf{Competing Methods.} As there are no existing methods for this task, we compare the method against the ablated models -- our full model without $\mathcal{L}_\text{CDD}$ (Ours w/o $\mathcal{L}_\text{CDD}$) and without distance feature (Ours w/o $D_\text{fg}$). \\
\noindent\textbf{Results.} We show the quantitative result of the motion in Table~\ref{tab:quant_motion}, and the result of the renderings in  Table~\ref{tab:quant_motion_rend}. The quality of the proposed method performs favorably against the competing methods while not sacrificing diversity. The diversity of Ours w/o $\mathcal{L}_\text{CDD}$ is slightly higher than Ours, albeit we observe a much higher Floating Err. This indicates that the diversity comes from generating more floating motions which do  not respect the background geometry.

\subsection{Qualitative Evaluation}
We show qualitative results of 4D generation and video rendering.

\noindent\textbf{Different Texturing of Foreground and Background.} 
We show qualitative results of different textured results in Fig.~\ref{fig:4.add_texture}. Given a novel mesh of foreground and background, we use Text2tex~\cite{text2tex} and different prompts to generate diverse textures for the meshes.

\noindent\textbf{Diverse Rendering of Videos.} 
We show qualitative results of diverse motion generation in Fig.~\ref{fig:4.render_ours}. In the top row, the motion model generates diverse motions by taking a different starting pose $G_0$, and it generates diverse motion trajectories such as jumping from to table to the sofa, or jumping down to the ground from the table. In the bottom row, the motion model also generates different motions with the same starting pose $G_0$ and different background $H_\text{bg}$ as input. For instance, jumping down from the bed or staying in the same position. The rendering results show that the proposed method can generate an environment-aware motion with temporal coherence in the unseen environment. 

\noindent\textbf{Comparison with Baseline Methods.} 
We show a qualitative comparison with competing methods in Fig.~\ref{fig:4.render_comp}. Our full model generates the motion sequence which is the most compatible with the background. For example, in Scene 2, the cat in our result jumps down from the bed successfully, while the results of other methods depict a cat that is either moving disregarding scene geometry, or floating unnaturally.

\begin{table}[t]
\caption{
    \textbf{Quantitative evaluation of motion}.
    We evaluate the trajectory reconstruction ability, motion diversity, floating artifacts.
}
\vspace{\tabmargin}
\vspace{-4mm}
\centering
\setlength{\tabcolsep}{0.25em}

\begin{tabular}{l cccc} 
    \toprule
    {Method} &  {Recon. $\shortdownarrow$ }  & {Diversity $\shortuparrow$ } & {Floating Err. $\shortdownarrow$ } \\
    \midrule
    Ours w/o $\mathcal{L}_\text{CDD}$   & 110.5709 & \textbf{103.9038} & 45.6105 \\
    Ours w/o $D_\text{fg}$  & 108.1872  & 87.1041 & 43.4656 \\
    Ours     & \textbf{103.4322} & 100.9871 & \textbf{38.1343} \\
    \bottomrule
\end{tabular}
\vspace{\tabmargin}
\vspace{-.5em}
\label{tab:quant_motion}
\end{table}

\begin{table}[t]
\caption{
    \textbf{Quantitative evaluation of  video rendering}.
    We evaluate the visual fidelity of the rendered videos. 
}
\vspace{\tabmargin}
\vspace{-3mm}
\centering
\setlength{\tabcolsep}{0.25em}

\begin{tabular}{l cccc} 
    \toprule
    {Method} &  {FID $\shortdownarrow$ } \\
    \midrule
    Ours w/o $\mathcal{L}_\text{CDD}$   & 125.3224 \\
    Ours w/o $D_\text{fg}$ & 126.4011 \\
    Ours     & \textbf{120.4663} \\
    \bottomrule
\end{tabular}
\vspace{\tabmargin}
\vspace{-1.5em}
\label{tab:quant_motion_rend}
\end{table}

\section{Conclusion and Discussion}
\label{6.conclusion}

We introduce Virtual Pet, a novel framework designed to acquire an environment-aware generative model for 3D motions from a monocular video collection. To extract shape, motion, and affordance from a set of cat videos, we propose a two-stage reconstruction strategy utilizing a combination of a deformable NeRF for articulated 3D reconstruction, and a static NeRF for background scene reconstruction. With the extracted data, we then adopt VAEs to learn to model the motion sequence, consisting of a trajectory generation and articulation generation. The generated motions are environment-aware, incorporating considerations for both foreground and background shapes.

To generate the 4D output, the proposed method operates on a foreground mesh and a background mesh.
The generated 4D output can then be rendered into videos given a camera pose and a renderer. 

Despite the achieved promising results, our current pipeline also has  limitations.
First, as the research of 4D reconstruction is still in the early stage, the quality of reconstructed motions has room for improvement.
Also, the limited robustness imposes constraints on the filtering of available videos.
Second, some generated motions deviate from physical rules, such as gravity.
Third, the model is currently species-specific.  

For future work, we aim to model finer details of the appearance and the motion. For instance, the fur for different species of the cat, and how it affects the motion. We also leave controllability of the output to future work, such as controlling the cat with natural language.
Furthermore, we envision augmenting our training data source with designed 3D assets as well as 2D prior knowledge from large-scale text-to-image models.

\noindent\textbf{Acknowledgements.} Work supported in part by NSF grants 2008387, 2045586, 2106825, MRI 1725729, and NIFA award 2020-67021-32799.

\clearpage
{
    \small
    \bibliographystyle{ieeenat_fullname}
    \bibliography{reference}
}

\end{document}